\documentclass[10pt, a4paper]{article}

\usepackage[final]{lrec2026} 
\usepackage{booktabs}
\title{Enhancing Naturalness in LLM-Generated Utterances\\ through Disfluency Insertion}

\name{Syed Zohaib Hassan$^{1}$, Pierre Lison$^{2}$, Pål Halvorsen$^{1}$} 

\address{$^{1}$SimulaMet, Oslo, Norway\\
         $^{2}$Norwegian Computing Center, Oslo, Norway\\
         \texttt{\{syed, paalh\}@simula.no, plison@nr.no}}

\abstract{
Disfluencies are a natural feature of spontaneous human speech but are typically absent from Large Language Models (LLMs) outputs. This absence can diminish the perceived naturalness of synthesized speech, which is an essential criterion when building conversational agents that aim to mimic human behaviors. We show how the insertion of synthetic disfluencies in the outputs can alleviate this shortcoming. The proposed approach involves (1) fine-tuning an LLM with Low-Rank Adaptation (LoRA) to incorporate various types of disfluencies into LLM-generated utterances and (2) synthesizing those utterances using a text-to-speech model that supports the generation of speech phenomena such as disfluencies. We assess the quality of the disfluency insertion through the use of automated metrics combined with a human study evaluating the disfluency insertion mechanism across two dimensions: \emph{intelligibility} and \emph{perceived spontaneity}. Token-level evaluation revealed that the model successfully learned patterns of disfluency placement but showed a tendency toward over-generation. Furthermore, the user study demonstrated that the insertion of disfluencies significantly increases the perceived spontaneity of the generated speech. This increase came, however, along with a slight reduction in intelligibility. 
\\ \newline  
}

\begin{document}

\maketitleabstract

\section{Introduction}

Disfluencies are an intrinsic component of spontaneous speech, often manifesting as hesitations, fillers ('uh,' 'like'), or repeated words and phrases. They play an essential role in human speech communication and are common across all types of non-scripted dialogues~\cite{lickley1998can, yaruss2002academic}. Pauses and hesitations often serve as markers of spontaneous thought \citep{saryazdi2021use}, and previous studies have highlighted that disfluent speech is often perceived as more spontaneous and natural~\cite{kampf2022perceived}.

This perception of naturalness by human listeners is important when building conversational systems designed to emulate human behaviors in various settings. Such systems are often described as virtual \textit{avatars} and are employed in applications such as educational tools, gaming, and healthcare support~\cite{chheang2024towards,qin2023empowering,yan2024general}. Previous studies have shown that factors like appearance, speech, lip-sync, and a strong sense of presence are essential for user engagement and effectiveness with these avatars~\cite{hassan2022towards,salehi2022more}, along with the quality of speech synthesis \cite{mattheyses2015audiovisual}. 

Avatars have also been employed to train human professionals to handle sensitive dialogues, such as delivering bad news in healthcare settings \citep{andrade2010avatar} or conducting investigative interviews with children who have experienced abuse~\cite{pompedda2015simulations,dalli2021technological,baugerud2021multimodal}. As emotional stress tends to decrease speech fluency \citep{buchanan2014acute}, disfluencies are highly frequent in those stressful and challenging dialogues, and an avatar developed for such training purposes should seek to reproduce such conversational behaviors. 

This paper presents a new method for achieving this objective. More specifically, the proposed approach artificially inserts disfluencies in non-disfluent utterances and passes on the results to a speech synthesis model that is well-suited for generating spontaneous speech. The disfluency insertion is performed by an LLM specifically fine-tuned for this task using LoRA on the basis of the Switchboard corpus \citep{godfrey1992switchboard}.

This paper presents three contributions: 
\begin{itemize}
\item The fine-tuning of an open-weight LLM to introduce disfluencies in dialogue utterances, which are then used as input to a text-to-speech model. 
\item The evaluation of the quality of the disfluency insertion through reference-based metrics. 
\item A user study to evaluate the naturalness and intelligibility of the disfluent speech generated by such a process.
\end{itemize}

Section \ref{sec:background} provides a brief overview over existing work on disfluencies and disfluency generation. Section \ref{sec:methods} presents the approach developed to automatically insert disfluencies in non-disfluent utterances. This approach is then evaluated by comparing the resulting utterances to the ones found in Switchboard (Section \ref{sec:results}) and by conducting a user study (Section \ref{sec:user_study}). Finally, Section \ref{sec:discussion}  discusses those empirical results, and Section \ref{sec:conclusion} concludes the paper. 

\section{Background}
\label{sec:background}

\citet{shriberg1994preliminaries} demonstrated that speech disfluencies are not random but follow specific patterns. They can be categorized as typical (e.g., hesitations, fillers like "um" and "uh") or atypical (e.g., repetitions, substitutions, insertions, and speech errors~\cite{yaruss2002academic}). Table~\ref{tab:disfluency_classes} shows examples of different types of disfluencies commonly observed in spontaneous speech.



\begin{table}[t]
    \centering\footnotesize
    \begin{tabularx}{\linewidth}{lX}
        \toprule
        \textbf{Disfluency type} & \textbf{Example} \\ \midrule
         
        Repetition                & \st{\textbf{\textit{he}}} he didn't call me\\
        Substitution              & \st{\textbf{\textit{she went to}}} he went to the store\\
        Filled Pause              & I saw \st{\textbf{\textit{uh}}} the movie yesterday\\
        Insertion                 & \st{\textbf{\textit{well}}} I was thinking of going there\\  
        Speech Error              & I remember your \st{\textbf{\textit{bir-}}} birthday.\\\bottomrule
        
    \end{tabularx}
    \caption{Examples of speech disfluencies. Strike-through formatted words represent the parts that should be edited out to obtain a fluent utterance, and correspond to the \textit{reparandum} and \textit{interregnun} in \citet{shriberg1994preliminaries}'s terminology.}\label{tab:disfluency_classes}
\end{table}

A few previous works have explored the generation of disfluencies in spoken utterances. To our knowledge, \citet{marge2010towards} presented the first system to insert disfluencies in dialogue system outputs, relying on heuristic rules instead of a fine-tuned LLM. \citet{passali2022lard} introduced large-scale artificial disfluency (LARD), which generates disfluencies like repetitions, replacements, and restarts using a synthetic approach that selects random positions within fluent text sequences, modifying them algorithmically to create disfluent variants.


\citet{yang-etal-2020-planning} introduced a Planner-Generator model to generate diverse disfluent texts for training disfluency detection models. The Planner identifies optimal disfluency insertion points, while the Generator produces corresponding phrases.


\citet{marie-2023-disfluency} fine-tuned the base T5 model~\cite{raffel2020exploring} for few-shot learning on the Fisher corpus, leveraging fluent-disfluent parallel subsets of varying sizes and the complete set of utterances. This paper builds on their approach and extends it along several dimensions. In addition to using a larger, decoder-only model for the disfluency generation, we also provide a detailed human evaluation of how the insertion of disfluencies affects their perceived spontaneity and intelligibility. 


\section{Methods}
\label{sec:methods}

We now describe the fine-tuning approach employed to insert disfluencies in non-disfluent utterances, as well as the text synthesis model converting those into speech. 


\subsection{Dataset}
We used the Switchboard dataset with disfluency annotations by~\citet{zayats2019disfluencies}. The dataset was based on the original Switchboard corpus~\cite{godfrey1992switchboard}, previously annotated by~\citet{marcus1999treebank} and subsequently cleaned and re-annotated using BIO tagging to accurately capture reparandum and correction spans.
We also used the NXT Switchboard Corpus~\cite{calhoun2010nxt} to enrich the aforementioned data with explicit annotations of silent pauses, which often signal hesitation, planning, or shifts in dialogue.


We represent the training data as pairs of utterances $(u_f, u_d)$ where $u_d$ is the transcription of the original utterance (thus including disfluencies) in the Switchboard corpus, and $u_f$ corresponds to its non-disfluent equivalent after the removal of all disfluencies from the transcription. The data from~\citet{zayats2019disfluencies} was used to extract instances of atypical disfluencies, while~\citet{calhoun2010nxt} provided instances of typical disfluencies. Table~\ref{tab:stats} shows an overview of both the training and test sets. The rate of disfluency is defined as the ratio of disfluent tokens to the total number of tokens in an utterance. 

\begin{table}[t]
    \centering\footnotesize
    \begin{tabularx}{\linewidth}{@{\hspace{0mm}}l@{\hspace{1mm}}XX@{\hspace{0mm}}}
        \toprule
        \textbf{Dataset} &  \textbf{Train} & \textbf{Test}\\
        \midrule
        No. Sentences & 32490& 3610 \\
        Avg No. of tokens in fluent utterance & 24.28 & 24.15 \\
        Avg No. of tokens in disfluent utterance & 33.08 & 32.84 \\
        Total No. fluent tokens & 789K & 87K \\
        Total No. disfluent tokens  & 1075K & 119K\\
        Rate of disfluency (\%)  & 24.5\% & 23.9\% \\\bottomrule
    \end{tabularx}
    \vspace{-1mm}
    \caption{Statistics of the training set used for fine-tuning the LLM and the test set used for evaluation.}\label{tab:stats}
    \vspace{-3mm}
\end{table}

\subsection{Model Training}

We fine-tuned Llama-2-7b-chat-hf~\cite{touvron2023llama} and Flan-T5-Large ~\cite{chung2024scaling} with LoRA~\cite{hu2021lora} to generate disfluent utterances from fluent inputs. The fine-tuning setup included a maximum sequence length of 200 tokens to balance accuracy with computational efficiency. We selected LoRA for its computational efficiency and parameter-effective approach, which updates only a small subset of rank-specific parameters rather than the entire model as in supervised fine-tuning (SFT). This minimized over-fitting risks while enabling targeted adaptations. To balance generalization and efficiency, we set LoRA rank (r) to 32, scaling factor (alpha) to 64, and dropout rate to 0.1. Training employed a batch size of 2 with gradient accumulation steps of 4, simulating larger batches on limited resources, and a learning rate of $2\times10^{-4}$. The model was fine-tuned for two specific tasks: generating atypical disfluencies (repetition, substitution, insertions, speech pauses) and typical disfluencies (filled and silent pauses).


\subsection{Text-to-speech}

Disfluencies are a speech phenomenon. We use existing text-to-speech models to convert the disfluent utterances generated by the fine-tuned LLM into audio form. To select a TTS model for generating realistic disfluent speech, we conducted a pilot study in which we evaluated audio samples generated by three TTS models: Bark TTS model developed by Suno-AI~\cite{bark:github},
Tortoise TTS~\cite{betker2023better} and OpenAI TTS~\cite{openai:tts}.
The evaluation focused on the prosody, intonation, and, in particular, the pronunciation of disfluencies. An important consideration was how the text-to-speech rendered false starts such as "Th- they" or "B- birthday", which are particularly frequent in Switchboard -- and hence in the disfluent utterances produced by the fine-tuned LLM. 

Among the three models, Bark TTS was deemed the most effective at synthesizing these disfluencies into speech. While all three TTS models sometimes produced errors, such as buzzing sounds and prolonged disruptive pauses in longer utterances, we found that Bark TTS provided the most natural-sounding speech when applied to our disfluent transcriptions. To ensure the audio clips generated with Bark for the user study were high quality, we regenerated samples when noisy artifacts, such as buzzing sounds, were detected in the audio outputs. 


\section{Evaluation}
\label{sec:results}

To assess the performance of our disfluency insertion approach, we start with automatic evaluations of the fine-tuned models. More precisely, we examine the extent to which those models replicate disfluency patterns found in natural speech, by comparing overlaps between real disfluent utterances and their synthetically generated counterparts.




\subsection{Reference-based Metrics}

We first compare the disfluent utterances generated by each fine-tuned model with the actual utterances in the Switchboard test set. We used BLEU and BERTScore as metrics: while BLEU \cite{papineni2001method} measures the lexical overlap between the generated outputs and the reference transcriptions, the BERTScores \cite{zhang2019bertscore} indicates the model’s ability to maintain semantic consistency between the generated and reference texts. Table~\ref{tab:results} presents the results. The \textrm{llama-2-7B} model achieved a higher BLEU score, indicating better lexical overlap with reference transcriptions. Conversely, \textrm{flan-t5} showed stronger semantic precision but lower recall, resulting in a slightly higher overall $F_1$ score. 

\begin{table}[t]
    \centering\footnotesize
    \begin{tabular}{lcc}
        \toprule
        \textbf{Metric} & \textbf{llama-2} & \textbf{flan-t5} \\
        \midrule
            BLEU Score & \textbf{0.5524} & 0.4440 \\
            BERTScore Precision & 0.9287 & \textbf{0.9624} \\
            BERTScore Recall & \textbf{0.9370} & 0.9168 \\
            BERTScore F1 & 0.9327 & \textbf{0.9386} \\
        \bottomrule
    \end{tabular}
    \caption{Performance of two fine-tuned LLMs in reproducing the disfluency patterns found in the Switchboard test set, measured using BLEU and BERTScore.}\label{tab:results} \vspace{-2mm}
\end{table}




As \textrm{llama-2-7b} produced disfluent utterances that exhibited a higher overlap (based on BLEU) with human-produced disfluent utterances, we conducted a token-level evaluation for disfluency placement exclusively on this model to understand not only whether it generates appropriate disfluencies, but also where it places them relative to human-produced disfluencies, as explained below. 

\subsection{Disfluency placement}

\begin{table}[t]
\centering\footnotesize

\begin{tabular}{lcc}
\toprule
\textbf{Metric} & \textbf{Value} & \textbf{95\% CI} \\
\midrule
Precision & 52.2\% & [51.4, 53.0] \\
Recall & 85.8\% & [85.0, 86.6] \\
F1-Score & 61.2\% & [61.0, 61.9] \\
Accuracy & 81.6\% & [81.4, 81.8] \\
Jaccard Similarity & 30.4\% & [29.8, 31.0] \\
AUC Score & 0.603 & \\
\bottomrule
\end{tabular}
\caption{Token-level overlap between human-produced disfluencies and disfluencies synthetically generated with our model ($N$=3,610 utterances, for a total of 96,573 tokens).}
\label{tab:positional_metrics}

\end{table}

To determine whether the fine-tuned LLM (llama-2-7b) mimics disfluencies in the same locations and types as humans, we then compared the relative positions of disfluencies in human and LLM-generated utterances using standard token-level evaluation metrics (precision, recall, F1 score, and AUC score). We manually annotated 3,610 fluent-disfluent sentence pairs from our test set and utterances with LLM-generated disfluencies using a rule-based approach, marking each token position as either disfluent (D tag) or fluent (O tag) in both the human-produced disfluent utterances and the model-generated outputs. Tokens were marked as disfluent if they were filler words (e.g., "uh", "um"), cut-off words (ending with "-"), absent from the original fluent text, or repetitions of previous words; otherwise, they were tagged as fluent.  

Table~\ref{tab:positional_metrics} presents the overall results of this comparison for token-level disfluency matching. Compared to the manually annotated disfluencies, the utterances with synthetic disfluencies generated by our approach demonstrated high recall but moderate precision, revealing a tendency towards over-generation. The Jaccard similarity coefficient averaged 30.4\% (±27.5\%), indicating moderate overlap between model predictions and human annotations.


\subsection{Type-Specific Accuracy}

The overlaps between human-produced and synthetic disfluencies varied significantly between disfluency types. Table~\ref{tab:type_specific_performance} factors this variation across typical disfluency insertions (SIL) and atypical disfluencies (DISFL). Typical disfluency insertions showed higher recall but lower precision than atypical disfluencies. The fine-tuned disfluency generation model generated substantially more typical disfluencies than humans ($12.9$ vs $5.0$ per sample on average), while atypical disfluencies showed more controlled generation patterns ($5.0$ vs $4.2$ per sample). 

\begin{table}[ht]
\centering\footnotesize

\begin{tabular}{lccc}
\toprule
\textbf{Metric} & \textbf{SIL} & \textbf{DISFL} & \textbf{Overall} \\
\midrule
\textbf{No. Samples  } & 1,094 & 2,516 & 3,610 \\
\textbf{Precision (\%)} & 49.5 & 53.4 & 52.2 \\
\textbf{Recall (\%)} & 93.4 & 82.5 & 85.8 \\
\textbf{F1-Score (\%)} & 61.6 & 61.1 & 61.2 \\
\textbf{Avg Disfl/Sample} & 12.9 & 5.0 & 7.4 \\
\midrule
\textbf{Human Baseline} & \multicolumn{3}{c}{4.2 Avg Disfl/Sample} \\
\bottomrule
\end{tabular}
\caption{Token-level overlap factored by disfluency type: silence insertions (SIL) or structural disfluencies (DISFL).}
\label{tab:type_specific_performance}
\end{table}





\section{User Study}
\label{sec:user_study}

To assess how the insertion of disfluencies in utterances affected their perception by human listeners, we conducted a user evaluation study comparing fluent and disfluent speech conditions on two criterias: \textit{intelligibility} and \textit{perceived spontaneity}. We designed these conditions to simulate real-life conversational scenarios, allowing us to observe how variations in speech fluency affect the listener's experience.

\subsection{Study Design}

\begin{figure*}[t]
     \centering
     \includegraphics[width=\linewidth]{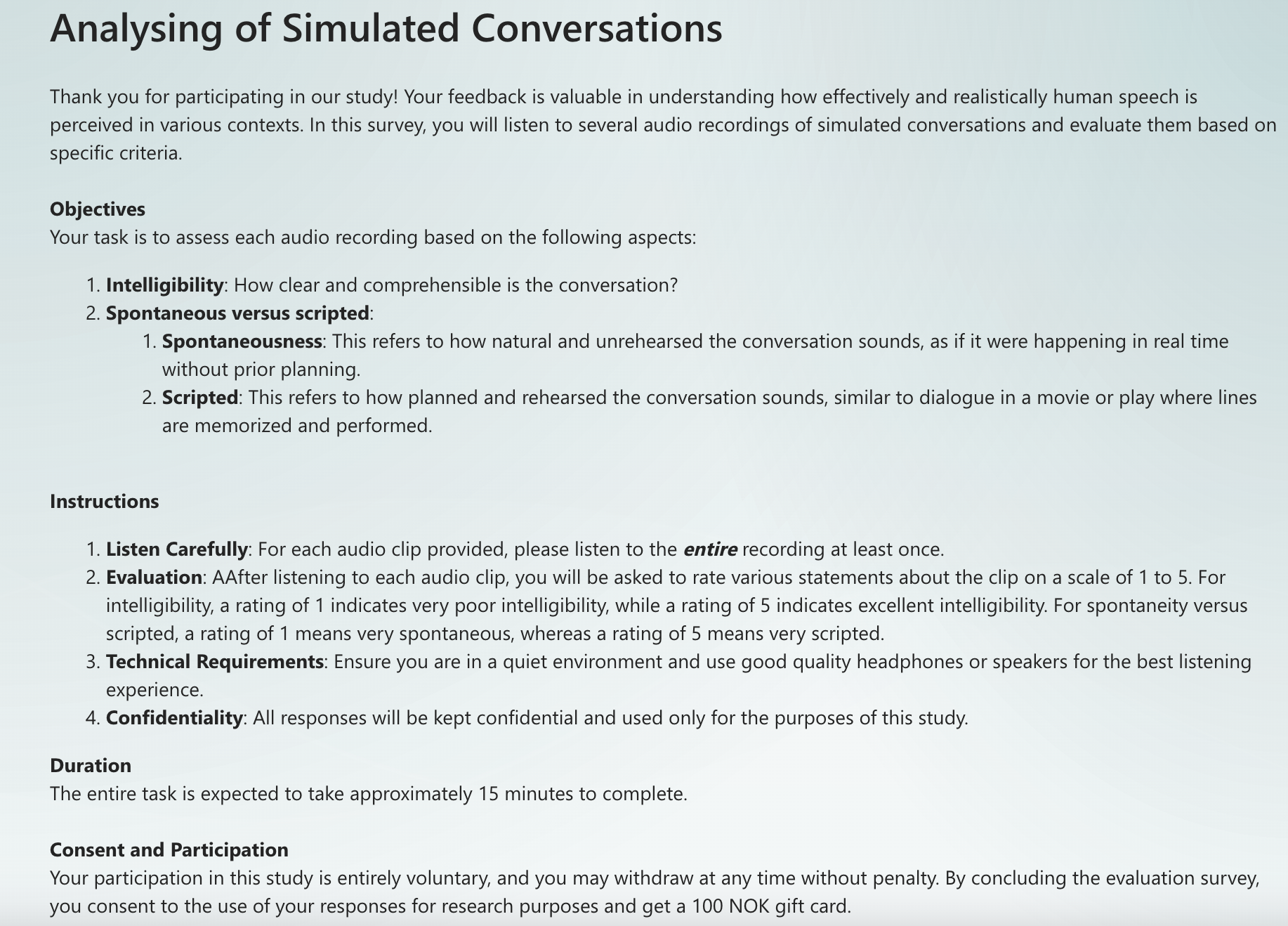}
     \caption{Instructions to the user study's participants}
     \label{fig:instructions}
\end{figure*}

In our user study, we asked participants to listen to 10 audio clips of simulated conversations, five clips each for disfluent and fluent audio. We used separate sets of conversation content to distinguish between the disfluent and fluent audio conditions. We used ChatGPT to generate ten fictive conversations based on real-life scenarios for human evaluation services, each with five turns of dialogue. We inserted disfluencies into five conversations using our fine-tuned language model, leaving the remaining five fluent. We converted the fluent and disfluent conversations into audio clips using the BARK TTS model. The instructions given to the participants are shown in Figure \ref{fig:instructions}, along with the transcriptions of the (respectively fluent and disfluent) dialogues to assess in Tables \ref{tab:study-data} and \ref{tab:study-data-2}. 

Participants were requested to assess each audio recording based on two criteria: 
\begin{itemize}

\item \textbf{Intelligibility}: How clear and comprehensible was the spoken conversation.
\item \textbf{Spontaneous versus scripted}: Did the conversation sound natural and unrehearsed, as if it were happening in real-time without prior planning? Or did the dialogue sound scripted in advance, similar to dialogue in a movie or play where lines are memorized and performed? 
\end{itemize}

\begin{table*}[thb]
\footnotesize
\begin{tabular}{p{0.16\linewidth}|p{0.78\linewidth}}
\hline
\textbf{Conversation} & \textbf{Conversation Data} \\ \hline
{\textbf{Conversation 1}} &
Speaker 1: Our solution will significantly improve your efficiency. \\& Speaker 2: I'm not sure if this fits within our budget. \\
&Speaker 1: We offer a proven track record of success. \\& Speaker 2: Can you provide some case studies? \\
&Speaker 1: This investment will pay off in the long run. \\ &Speaker 2: What is the expected ROI? \\
&Speaker 1: Our team will provide full support during implementation. \\ &Speaker 2: What kind of support do you offer? \\
&Speaker 1: Let's move forward with this proposal. \\ &Speaker 2: I'll need to discuss this with my team. \\
\hline
{\textbf{Conversation 2}} &
Speaker 1: It's been ages since we last met! \\ & Speaker 2: I know, time flies! \\
&Speaker 1: How have you been? \\ &Speaker 2: I've been good, just busy with work. \\
&Speaker 1: We should plan a trip together.\\ &Speaker 2: That sounds like a great idea, where should we go? \\
&Speaker 1: Do you remember our school days? \\ &Speaker 2: Of course, those were some of the best days. \\
&Speaker 1: Let's not wait too long to catch up again. \\ &Speaker 2: Absolutely, let's make it a regular thing. \\
\hline
{\textbf{Conversation 3}} &
Speaker 1: We need to finalize this deal soon. \\ &Speaker 2: We're also keen to close this deal promptly. \\ &
Speaker 1: Our terms are quite clear and non-negotiable. \\ &Speaker 2: We can discuss some flexibility in certain areas. \\ &
Speaker 1: We expect a long-term partnership. \\ &Speaker 2: We're committed to a sustainable partnership. \\ &
Speaker 1: Can we agree on a mutual benefit clause? \\ &Speaker 2: Yes, that seems reasonable and beneficial. \\ &
Speaker 1: What is your final offer? \\ &Speaker 2: Here's our final offer, let's seal the deal. \\
\hline
{\textbf{Conversation 4}} &
Speaker 1: Our goal is to inspire change and innovation. \\ &Speaker 2: We need to provide concrete data and examples. \\&
Speaker 1: We must focus on the big picture. \\ &Speaker 2: Details matter when explaining our strategy. \\&
Speaker 1: Every challenge is an opportunity. \\ &Speaker 2: What are the specific challenges and solutions? \\&
Speaker 1: Think about the impact we can make. \\ &Speaker 2: We should quantify the potential impact. \\&
Speaker 1: Let's engage the audience with our vision. \\ &Speaker 2: Our presentation should balance vision with facts. \\
\hline
\textbf{Conversation 5} &
Speaker 1: We need to complete this project by Friday. \\ &Speaker 2: Understood, I'll prioritize my tasks accordingly. \\&
Speaker 1: Everyone must focus on their assigned tasks. \\ &Speaker 2: I'll make sure to stay on track. \\&
Speaker 1: Report your progress at the end of each day. \\ &Speaker 2: I'll update you with my progress daily. \\&
Speaker 1: Ensure all deliverables meet our quality standards. \\& Speaker 2: I'll double-check my work for quality. \\&
Speaker 1: Let's work together to meet our deadline. \\ &Speaker 2: I'll collaborate with the team to ensure success. \\
\hline
\end{tabular}\caption{Fluent conversational data used in the user study}\label{tab:study-data}
\end{table*}

\begin{table*}[]
\footnotesize
\begin{tabular}{p{0.16\linewidth}|p{0.78\linewidth}}
\hline
\textbf{Conversation} & \textbf{Conversation Data} \\ \hline
{\textbf{Conversation 1}} & Speaker 1: I'm i'm um worried about um m um my child's perf um  performance in math. \\  &Speaker 2: I understand your concern and let's discuss a a plan. \\ &
Speaker 1: What  what can we do um  to improve the erm  their grades? \\  &Speaker 2: We can provide addi additional practice material and t tutoring sessions. \\ &
Speaker 1: Are there any additional um  re um  resources we can use at um at home? \\  &Speaker 2: Yes, I I'll send you those some online resources that can can be very helpful. \\ &
Speaker 1: Is my child erm  is participating in class? \\  &Speaker 2: Your child is quite en engaged, but we can en encourage more participation. \\ &
Speaker 1: How can we erm  how can we work together erm to um to support my ch um child's learning? \\  &Speaker 2: Communication is key let's set up reg regular updates and support. \\ 
\hline
{\textbf{Conversation 2}} &
Speaker 1: I've been um ha um having these headaches for for weeks. \\  &Speaker 2: Let's let's go through your your symptoms in in detail first. \\ &
Speaker 1: Could it be um could it be something serious um? \\  &Speaker 2: It's it's too early to to say, but but we'll we'll investigate thoroughly \\ &
Speaker 1: Should I get some erm some tests done um.\\  &Speaker 2: Yes, we we'll start with with a a few basic tests and and go from there. \\ &
Speaker 1: What  what can I do to to alleviate the  the pain? \\  &Speaker 2: I'll give you some medication and I'll suggest some lifestyle changes. \\ &
Speaker 1: Is there a specific um rea reason for these erm these symptoms? \\  &Speaker 2: It could be due stress or other factors but we'll find find out \\ 
\hline
{\textbf{Conversation 3}} &
Speaker 1: Can you um can you erm tell me about your um your previous work experience? \\  &Speaker 2: I've I've worked in in similar roles for f five years with with great results. \\ &
Speaker 1: What makes you a um a good fit for for this role.\\  &Speaker 2: I I have the skills and and passion that align with with this company's val values. \\ &
Speaker 1: How do you um  how do you handle tight erm  tight deadlines? \\  &Speaker 2: I prioritize prioritize tasks and and stay focused under under pressure. \\ &
Speaker 1: Why do you want erm why do you want to to work with us? \\  &Speaker 2: I admire this this company's vision a and want to to contribute to it. \\ &
Speaker 1: Where erm where do you see yourself in um in five years. \\  &Speaker 2: I see myself growing with this company and growing within this company and and taking on more respons responsibilities.\\ 
\hline
{\textbf{Conversation 4}} &
Speaker 1: How do I im um improve my leadership skills? \\  &Speaker 2: Take on m more responsibilities and and seek feedback regularly. \\ &
Speaker 1: What what should I focus on in in my career development. \\  &Speaker 2: Focus on learning continuous learning and networking. \\ &
Speaker 1: Can you erm can you give me feedback on erm on my recent project.\\  &Speaker 2: You did well but there's you're there's room for improv improvement in communication. \\ &
Speaker 1: How do you handle erm con erm workplace conflicts? \\  &Speaker 2: Approach conflicts with with a a calm and solution oriented mindset.\\ &
Speaker 1: What's the what's the erm best advice y erm  you've received in in your career? \\  &Speaker 2: Always be open to to learning and and never be afraid to to ask questions. \\ 
\hline
\textbf{Conversation 5} &
Speaker 1: Have you um have you reviewed the the latest project report? \\  &Speaker 2: Yes, I- I think we need a a more innova- innovative approach. \\ &
Speaker 1: The data suggests that a more traditional method is um is effective um.\\  &Speaker 2: But creativity could g- give us an edge.\\ &
Speaker 1: Are there any erm any risks associated with erm you- um your approach ? \\  &Speaker 2: Some some, but nothing we can n't handle. \\ &
Speaker 1: We need to erm con- we need to consider the erm the budget constraints as well erm \\  &Speaker 2: I- I believe we can we can be cost- effective and creative.\\ &
Speaker 1: I'll need to see a to see a detailed plan before.\\  &Speaker 2: I'll i'll draft something and and send it over. \\ 
\hline
\end{tabular}\caption{Disfluent conversational data used in the user study}\label{tab:study-data-2}
\end{table*}


We asked study participants to rate each clip on intelligibility and spontaneity on a 5-point Likert scale (from worst to best). 

The study involved 41 participants, with 21 aged between 25-34 years, followed by 13 participants aged 18-24 years, 6 participants aged 35-44 years, and one participant aged 45-54 years. They were asked about hearing loss, with 39 participants reporting no hearing loss, and 2 participants preferring not to answer. They accessed the user study through an online questionnaire filled from either their mobile phones (17), laptops (15), tablets (6), or desktop computers (3). 32 participants used headphones. Participation was voluntary, and each participant received a gift card as compensation. 


\subsection{Study Results}

We investigated the impact of inserting disfluencies into originally fluent spoken utterances, using human evaluations to compare intelligibility and spontaneity ratings. Due to its higher BLEU score, we used the fine-tuned llama-2-7b model to generate the disfluent utterances for this study.

\begin{figure}[t]
     \centering
     \includegraphics[width=\linewidth]{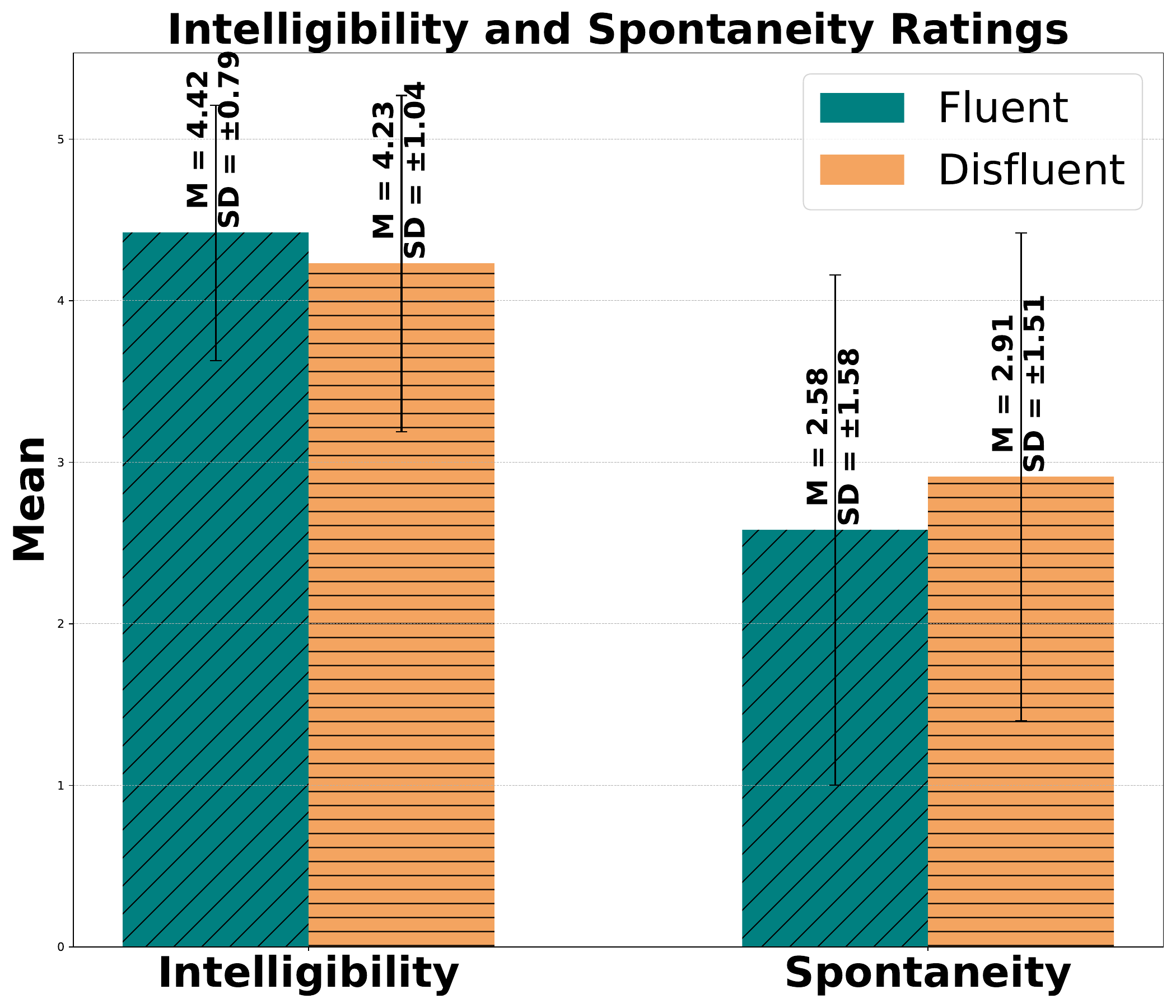}
     \caption{Bar-plot (mean values and 95\% confidence intervals) of user ratings on intelligibility and spontaneity.}
     \label{fig:stats}
 \end{figure}

As shown in Figure~\ref{fig:stats}, the average intelligibility ratings from the participants were \(4.42 \pm 0.79\) for fluent utterances and \(4.23 \pm 1.04\) for disfluent utterances. Spontaneity scored an average of \(2.58 \pm 1.58\) for fluent utterances and \(2.91 \pm 1.51\) for disfluent utterances. Fluent utterances demonstrated higher intelligibility and lower spontaneity compared to disfluent utterances. 

We assessed the statistical significance of those results using both independent two-sample T-tests and mixed-effects models, with results shown in Table~\ref{tab:results2}. For intelligibility, the $t$-statistic was $2.03$ ($p$ = $0.043$), and for spontaneity, the $t$-statistic was $-2.14$ ($p$ = $0.033$). Both $p$-values, below the $\alpha$ level of $0.05$, indicate statistically significant differences. As participants rated multiple clips in both conditions, we derived a mixed-effect analysis to account for within-subject variability and provide more accurate estimates of condition effects. For intelligibility, the mixed-effects model yielded a z-value of $2.809$ (p = $0.005$), confirming a statistically significant effect of the condition, consistent with t-test results. The group variance for intelligibility was $0.420$, indicating moderate variability across groups. For spontaneity, the model showed a z-value of $-2.615$ (p = $0.009$), again demonstrating a significant effect, with a group variance of $0.805$, reflecting greater between-group variability.

\section{Discussion}
\label{sec:discussion}

\begin{table}[t]
    \centering\footnotesize
    \begin{tabularx}{\columnwidth}{@{\hspace{0mm}}l@{\hspace{1mm}}l@{\hspace{1mm}}r@{\hspace{2mm}}rc@{\hspace{1mm}}r@{\hspace{0mm}}}
        \toprule
        \textbf{Metric} & \textbf{Test} & \textbf{Statistic} & \textbf{p-value} & \textbf{GV} \\ \midrule
        \textbf{Intelligibility} & T-Test & $t=2.03$ & $0.043$ & - \\ 
        & Mixed E.\ & $z=2.809$ & $0.005$ & $0.420$ \\ \midrule
        \textbf{Spontaneity} & T-Test & $t=-2.14$ & $0.033$ & - \\ 
        & Mixed E.\ \ \ \ & $z=-2.615$ & $0.009$ & $0.805$ \\ \bottomrule
    \end{tabularx}
    \caption{Statistical analysis of perceived intelligibility and spontaneity ratings. GV=Group Variance.}\label{tab:results2} \vspace{-2mm}
\end{table}

We sought to answer two questions: i) whether \textit{intelligibility decreases when we add the disfluencies to the fluent spoken utterances} and ii) whether \textit{perceived spontaneity increases when we add the disfluencies to the fluent spoken utterances}.

The main findings from this user study are that disfluencies significantly enhance the perceived spontaneity of speech while leading to a slight reduction in intelligibility. The first effect is, however, stronger than the second, as evidenced by the results in Table~\ref{tab:results2}. 
The statistical results obtained with the mixed-effect model also highlighted the variability within participants' responses. Greater variability in spontaneity ratings suggests that perceptions of speech realism are more subjective and shaped by individual expectations of what constitutes a spontaneous interaction. 

Our within-subjects design provided sufficient statistical power to detect medium-sized effects, validating our approach despite the relatively modest sample size. Audio clips used in the study ranged from approximately one minute for fluent samples to $1.5$ minutes for disfluent ones, providing participants adequate exposure to meaningfully perceive and evaluate the presence of disfluencies. While our current findings demonstrate the effectiveness of our fine-tuned LLM approach coupled with Bark TTS, future work could extend this research through larger-scale studies with expanded and diverse audio samples and participant pools to detect more subtle effects. Additionally, comparative evaluations with dialogue-optimized models like ChatTTS~\cite{chatTTS} could further enhance understanding of how different TTS systems render generated disfluencies in conversational contexts.

As shown in Table \ref{tab:results}, the fine-tuned LLMs effectively insert disfluencies while preserving the content of the original utterance. High BERTScore and BLEU scores confirm close alignment with reference disfluent utterances in the test set. Despite the sporadic nature of disfluencies in spontaneous speech, the model reliably mimics those in the Switchboard dataset. However, the average disfluency rate in the \textbf{llama-2-7b} generated disfluent text is 29.1\%, slightly exceeding the rate observed in the training data, as shown in Table~\ref{tab:stats}, which could be a focus in future work. 


The token-level evaluation reveals that the fine-tuned LLM is able to identify appropriate positions for disfluencies but unable to control their overall frequency. The high recall rate demonstrates that the model has successfully learned meaningful patterns of human disfluency placement, correctly identifying approximately 4 out of every 5 human-annotated locations. This ability to learn where disfluencies typically occur, combined with the 30.4\% Jaccard similarity, indicates that disfluencies follow learnable structural patterns that transformer architectures can effectively capture.
However, the systematic over-generation (78.1\% of samples) and 5:1 false positive ratio reveal that while the model knows \textit{where} to place disfluencies, it lacks restraint mechanisms governing \textit{when} to stop adding them. Further work is needed to explore how, e.g.~additional regularization or hyper-parameter tuning may be employed to mitigate this over-generation. 

The differential performance between disfluency types provides additional insights: atypical disfluencies showed better precision (53.4\%) and controlled generation, while typical disfluency insertions exhibited extreme over-generation. This pattern indicates that syntactic constraints provide clearer learning signals than the prosodic and pragmatic factors governing typical disfluency placement.

Our token-level evaluation employed binary labeling (disfluent vs. fluent) to assess whether the model generated the same specific type of disfluency as humans at matched positions. This limitation stems from the structure of available text-based disfluency corpora: while audio datasets like FluencyBank~\cite{ratner2018fluency} and SEP-28k~\cite{lea2021sep} provide categorical disfluency type labels, text-based datasets like Switchboard predominantly use BIO tagging schemes for span detection~\cite{zayats2019disfluencies,rocholl2021disfluency} rather than explicit types of disfluencies at the token level. Consequently, while we can assess aggregate type-specific performance (e.g., overall precision/recall for typical insertions vs. atypical disfluencies), we were not able to analyze disfluency type and positional correspondence, whether the model selects the same disfluency type as humans at matched positions.

These findings help explain our user study results, where disfluent speech increased perceived spontaneity but reduced intelligibility. The model's ability to learn appropriate placement patterns likely contributed to the enhanced spontaneity perception, while the 77\% over-production of disfluencies compared to natural human patterns may have negatively impacted intelligibility ratings. The results establish that current LLM architectures excel at learning structural dependencies but struggle with contextual appropriateness and pragmatic restraint—a critical consideration for applications requiring naturalistic speech modeling.


\section{Conclusion}
\label{sec:conclusion}

This paper showed that (1) disfluencies can be automatically inserted into non-disfluent utterances using a fine-tuned LLM, and (2) this insertion enhances the perceived spontaneity of synthetic speech. Our token-level evaluation revealed that the model successfully learns patterns of human disfluency placement but tends toward systematic over-generation, particularly for typical disfluency insertions. Our user study revealed a perceptual trade-off: participants rated utterances with inserted disfluencies as more spontaneous but slightly lower on the intelligibility scale.
These findings are particularly beneficial for designing LLM-powered conversational avatars, such as those employed to train human professionals to handle "difficult" and stressful dialogues, such as investigative interviews, in which disfluencies are particularly prevalent. Future work will focus on developing frequency calibration mechanisms to control disfluency rates for specific scenarios, such as simulating stress or emotional speech in sensitive conversations.

\section{Limitations}

The outputs of the fine-tuned LLM for inserting disfluencies has a slightly higher disfluency rate compared to the Switchboard training data, indicating a tendency to generate disfluencies slightly more often than in natural human speech. This discrepancy could affect the realism of synthesized conversations in certain contexts. While lexical and semantic alignment metrics (BLEU and BERTScore) demonstrated the fine-tuned model's effectiveness, these measures did not assess the quality of inserted disfluencies, such as their contextual appropriateness and placement.

Our token-level evaluation used binary labeling that distinguishes only between disfluent and fluent tokens without encoding specific disfluency types at each position. This methodological choice reflects a fundamental constraint in available text-based disfluency datasets. Due to a lack of text corpus with disfluency types annotations, we cannot assess whether the model replicates human disfluency type selection at matched positions—for instance, whether both human and model use a filled pause versus a repetition at the same location. Developing text-based annotations with token-level type labels would enable more granular analysis of whether models faithfully reproduce human disfluency patterns.

Although Bark TTS~\cite{bark2023} was the most effective TTS model for our study, it occasionally introduced artifacts like buzzing sounds or prolonged pauses, requiring manual intervention to regenerate samples in extreme cases. Additionally, the user study was limited to a small set of fictitious conversations generated by ChatGPT to simulate real-life scenarios. While representative, these scenarios may not fully capture the diversity of real-world conversational contexts where disfluencies occur. Both of these limitations could have influenced participants' perceptions during the study, potentially affecting the results.

\section{Ethical Considerations}
The use of language models to introduce disfluencies into fluent utterances presents some ethical challenges. One concern is the potential for deception: by mimicking human behavior, these models create speech that "sounds more human," potentially tricking the listener into believing they are interacting with a human speaker when they are not. Although our overarching objective is to develop avatars for professional training purposes in which participants are well aware that they are interacting with a virtual agent, deploying these models in contexts where users are not informed of the artificial nature of their interlocutor could lead to ethical concerns about trust, consent, and manipulation.
 
Disfluency patterns in generated utterances can also inadvertently reinforce harmful stereotypes about linguistic abilities across different demographic groups, particularly those with speech disorders, neurodivergent individuals, non-native speakers, and racial or ethnic minority populations. 

To mitigate these risks, we advocate for clear disclosure policies in any deployment of such models that simulate human speech. Users should be informed whenever they are interacting with an artificial agent, ensuring transparency and preserving trust. Additionally, we have released our code and aggregated data via GitHub to ensure reproducibility (anonymized URL). To mitigate risks, we provide usage guidelines prohibiting applications that could deceive users about agent identity or reinforce linguistic stereotypes. Our resources should only be used in contexts with informed consent and transparent disclosure of AI-generated speech.



\section{Bibliographical References}\label{sec:reference}

\bibliographystyle{lrec2026-natbib}
\bibliography{lrec2026-example}


\end{document}